\documentclass[10pt,twocolumn,letterpaper]{article}

\usepackage[pagenumbers]{cvpr} 

\definecolor{cvprblue}{rgb}{0.21,0.49,0.74}
\usepackage[pagebackref,breaklinks,colorlinks,allcolors=cvprblue]{hyperref}

\usepackage{graphicx}
\usepackage{amsmath}
\usepackage{amssymb}
\usepackage{xcolor}
\usepackage{booktabs}
\usepackage{multirow}
\usepackage{capt-of}
\usepackage{xspace}
\usepackage{colortbl}
\usepackage{xcolor}
\usepackage{textcomp}
\usepackage{algorithm}
\usepackage[accsupp]{axessibility}  

\def\methodname{GenAssets}

\def\paperID{8513} 
\def\confName{CVPR}
\def\confYear{2025}

\title{GenAssets: Generating in-the-wild 3D Assets in Latent Space}

\author{
Ze Yang$^{1, 2}$
\quad Jingkang Wang$^{1, 2}$
\quad Haowei Zhang$^{1}$ \\
\quad Sivabalan Manivasagam$^{1, 2}$
\quad Yun Chen$^{1, 2}$
\quad Raquel Urtasun$^{1, 2}$ \\
\normalsize{
$^{1}$Waabi
\quad $^{2}$University of Toronto} \\
{\small\texttt{\{zyang, jwang, hzhang, siva, ychen, urtasun\}@waabi.ai}}
}

\begin{document}

\twocolumn[{%
\renewcommand\twocolumn[1][]{#1}%
\maketitle
	\vspace{-10mm}
    \begin{center}
	\includegraphics[width=1.0\textwidth]{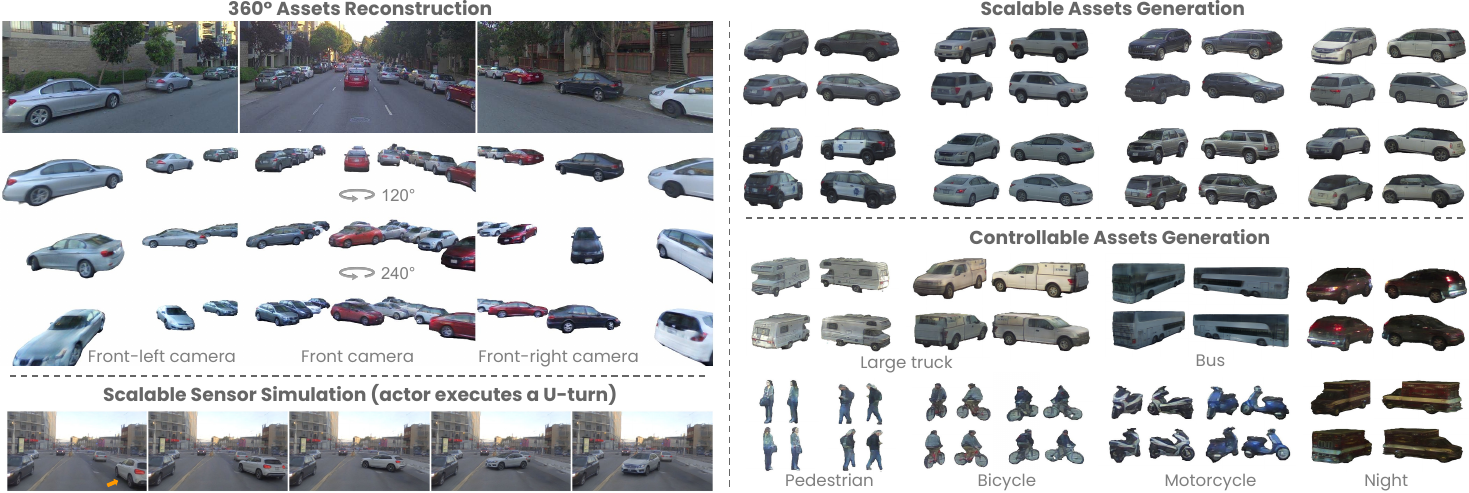}
    \end{center}
    \vspace{-6.2mm}
	\captionof{figure}{
	\textbf{\methodname{} takes in-the-wild camera image(s) and point cloud(s), and automatically reconstruct  or generate 360° assets}. Our 3D assets are diverse and high-quality with complete geometry and appearance, allowing for realistic and scalable sensor simulation.
	}
    \label{fig:teaser}
	\vspace{4mm}
}]

\begin{abstract}
    High-quality 3D assets for traffic participants are critical for multi-sensor simulation, which is essential for the safe end-to-end development of autonomy.
    Building assets from in-the-wild data is key for diversity and realism, but existing neural-rendering based reconstruction methods are slow and generate assets that render well only from viewpoints close to the original observations, limiting their usefulness in simulation.
    Recent diffusion-based generative models build complete and diverse assets, but perform poorly on in-the-wild driving scenes, where observed actors are captured under sparse and limited fields of view, and are partially occluded.
    In this work, we propose a 3D latent diffusion model that learns on in-the-wild LiDAR and camera data captured by a sensor platform and generates high-quality 3D assets with complete geometry and appearance.
    Key to our method is a ``reconstruct-then-generate'' approach that first leverages occlusion-aware neural rendering trained over multiple scenes to build a high-quality latent space for objects, and then trains a diffusion model that operates on the latent space.
    We show our method outperforms existing reconstruction and generation based methods, unlocking diverse and scalable content creation for simulation.
    Please visit \url{https://waabi.ai/genassets} ~for more details.
\end{abstract}
\vspace{-12pt}
\section{Introduction}
\label{sec:intro}

Simulation environments enable testing the performance of autonomy systems in long-tail and safety critical scenarios safely, efficiently, and scalably~\cite{unisim,wang2021advsim,manivasagam2023towards,sarva2023advd}.
To test the complete autonomy stack, the virtual environments should simulate the sensor data (\textit{e.g.}, LiDAR, camera) of the robot.
Realistic and consistent sensor simulation across multiple modalities depends on the availability of high-quality 3D \textit{assets} that accurately represent the geometry and appearance of traffic participants such as cars and motorcycles.

Commoditized simulation environments rely on artists to manually build such 3D assets \cite{dosovitskiy2017carla,shah2018airsim,Rong2020}.
This process is slow, costly, and requires precise specification of attributes such as geometry, UV mapping, and materials.
As a consequence, existing methods utilize very limited content, failing to capture the full diversity of objects in the real world.

An alternative and much more appealing approach that has gained traction is to reconstruct assets from real world data captured by the sensor platform, which enables diversity and realism.
Reconstruction-based neural rendering methods have demonstrated impressive results by optimizing 3D representations that can render and match input sensor data for self-driving scenes \cite{unisim, tonderski2024neurad, wu2023mars, tancik2022block, yan2024street, fischer2024dynamic, zhou2024drivinggaussian, khan2024autosplat, yang2023emernerf, chen2024omnire}.
These approaches render high-quality data when the viewpoint is close to the training views, but can suffer from severe artifacts at novel viewpoints due to incomplete observations.
Additionally, each asset must be observed in the real world and reconstructed via optimization, making it computationally expensive and limiting diversity.

Recent 3D generative models~\cite{niemeyer2021giraffe, chan2021pi, chan2022efficient, gao2022get3d, shen2023gina, roessle2024l3dg} have shown promise in generating complete assets and rendering novel views far from source.
These methods leverage a combination of generative adversarial networks (GANs), diffusion models, and neural rendering, and train on large corpuses of images to learn shape and appearance priors.
NeRF-based diffusion models that optimize via score-distillation sampling (SDS) \cite{poole2022dreamfusion, lin2023magic3d}, or denoise in 3D space \cite{muller2023diffrf, gu2023nerfdiff, chen2023single, wu2024direct3d, ntavelis2023autodecoding}, have demonstrated particularly high-quality complete shape and appearance generation, and also offer controllability through text/image conditioning.
However, such models usually rely on synthetic datasets with ground-truth 3D models to render the dense 2D views necessary to learn a 3D representation prior.
They also primarily work on ``clean'' images that fully capture the object of interest.
This reliance on clean synthetic data limits their ability to handle in-the-wild real data from a moving platform, where the data is typically captured from sparse viewpoints, under partial occlusions, limited resolution, and with sensor noise (\eg, lighting artifacts, rolling shutter).

In this work, we propose a latent diffusion 3D generative model that learns directly from sparse, in-the-wild data, enabling high-quality asset generation at scale.
We tackle the challenges of asset completion and generative modelling via a two-stage ``reconstruct-then-generate'' methodology.
In the first stage, we learn a low-dimensional object latent space that generates complete assets by training across multiple scenes via neural rendering.
To handle partial occlusion, we jointly learn a scene representation for both the static scene and dynamic actors, enabling occlusion-aware rendering during learning.
In the second stage, we train a diffusion model to operate on the 3D latent space, enabling realistic asset generation that can be conditioned on individual views, day or night, or the actor class, enabling asset variations that are applicable to new scenes.
We evaluate our model on the public self-driving dataset PandaSet \cite{xiao2021pandaset}, demonstrating its capability to reconstruct and generate high-quality assets from sparse, in-the-wild data.
Additionally, we showcase applications such as conditional generation, as well as realistic rendering of our assets for downstream simulation systems, highlighting  our approach's flexibility, scalability, and practical utility.

\section{Related Work}
\label{sec:related_work}

\paragraph{3D Reconstruction in the Wild:}
Reconstructing objects from in-the-wild data is essential for creating 3D digital twins and simulations, but poses significant challenges due to the complexity (\eg, occlusions, sparse viewpoints, partial observations) and noise (\eg, localization and calibration inaccuracies \cite{yang2025unical}, sensor noise).
Object-based reconstructions fall into optimizing explicit meshes \cite{munkberg2022extracting,wang2022cadsim,yang2021recovering,chen2021geosim} or implicit representations \cite{zhang2021ners,yang2023reconstructing,oechsle2021unisurf,wang2021neus,mildenhall2021nerf,jang2021codenerf,yang2021s3} methods via differentiable rendering.
However, they rely on accurate object segmentation masks and struggle to handle complex occlusions.
To address this issue, recent works propose compositional neural fields \cite{unisim,tonderski2024neurad,ost2021neural,pun2023neural,liu2023neural} or 3D Gaussians \cite{yan2024street,khan2024autosplat,chen2024omnire} to decompose 3D world into static background and dynamic actors.
These works often involve costly per-scene optimizations and cannot generalize well to novel views with large shifts nor produce complete (\ie, 360-degree) objects.
To mitigate these limitations, recent works propose generalizable reconstruction models \cite{muller2022autorf,chen2025g3r,hong2023lrm,wei2024meshlrm,ren2024scube} that generate a 3D representation with a single feed-forward network trained on many scenes.
Another line of work aims to incorporate data-driven priors from pre-trained models (\eg, stable diffusion \cite{rombach2022high}) via score-distillation sampling (SDS) \cite{poole2022dreamfusion,wu2024reconfusion,hwang2024vegs,wang2024freevs}.
However, existing methods either require object-centric synthetic training data \cite{hong2023lrm,wei2024meshlrm,wu2024reconfusion} (\eg, Objaverse~\cite{deitke2023objaverse}) and/or perform poorly on in-the-wild sensor data \cite{chen2025g3r,hwang2024vegs,wang2024freevs}.
Our approach builds high-fidelity and complete assets by directly learning on real-world data.

\paragraph{GAN-based 3D Generation:}
Generative Adversarial Networks \cite{goodfellow2014generative} (GANs) have shown great potential for 3D asset generation \cite{gao2022get3d,xu2023discoscene}, where the generator transforms random noise into the target 3D representation, and the discriminator is jointly trained using 2D images rendered from the generated representation and the ground truth.
This paradigm allows generation without requiring explicit 3D ground truth.
Specifically, $\pi$-GAN \cite{chan2021pi} introduces a generative model for high-quality 3D-aware image synthesis based on neural radiance fields.
GIRAFFE \cite{niemeyer2021giraffe} further proposes to represent scenes as compositional generative neural feature fields, allowing for better object-background disentanglement and controllable scene generation.
EG3D \cite{chan2022efficient} and GET3D \cite{gao2022get3d} further improve 3D generation efficiency and quality by integrating the triplane representation (for volume rendering or explicit mesh rendering) with a StyleGAN \cite{karras2019style} architecture.
Although they achieve robust and consistent 3D generation, they primarily focus on ``clean'' object-centric images with full coverage and no occlusions and have difficulties generalizing to in-the-wild driving data.
Most recently, DiscoScene \cite{xu2023discoscene} and GINA-3D \cite{shen2023gina} learn 3D generative models from in-the-wild data.
However, they often produce significant visual artifacts (\eg, floating particles, low-fidelity shapes and texture).
In contrast, our method generates high-quality assets.

\paragraph{Diffusion-based 3D Generation:}
Diffusion-based methods have achieved tremendous success in image generation \cite{ho2020denoising,song2020denoising,song2020score,rombach2022high,podell2024sdxl}, leveraging denoising diffusion probabilistic models \cite{ho2020denoising} to transform white noise into high-quality outputs through iterative refinement.
To bridge the gap between 2D and 3D generation, a common strategy involves lifting 2D images to 3D via score distillation sampling (SDS) \cite{poole2022dreamfusion, lin2023magic3d,wu2024reconfusion,hwang2024vegs,wang2024freevs,tang2023dreamgaussian,shi2023mvdream,gao2024cat3d}.
These methods rely on pre-trained or fine-tuned 2D diffusion models and optimize 3D representations with diffusion-prior guidance.
The optimization is slow and the supervision can be multi-view inconsistent.
To address this issue, recent works \cite{zero123,liu2023one2345,li2023instant3d,shi2023mvdream,long2024wonder3d,wang2023imagedream} propose to synthesize multi-view consistent images via diffusion models and then perform 3D reconstruction with explicit meshes \cite{xu2024instantmesh}, neural radiance fields \cite{li2023instant3d,wang2023imagedream} or 3D Gaussians \cite{tang2025lgm,xu2024grm}.
Another line of work performs diffusion directly in 3D, either by training 3D-aware diffusion model from posed 2D imges without requiring explicit 3D supervision \cite{wu2024neural,shue20233d,muller2023diffrf,gu2023nerfdiff,chen2023single,anciukevivcius2023renderdiffusion,cao2024lightplane,szymanowicz2023viewset,tewari2023diffusion,xu2023dmv3d} or by operating directly on 3D data \cite{luo2021diffusion,vahdat2022lion,liu2023meshdiffusion}.
Inspired by latent diffusion models (LDM) \cite{rombach2022high}, recent works diffuse within a latent space, where models like UNet \cite{ronneberger2015u} or VAEs \cite{kingma2013auto,van2017neural} map 3D data to a compressed latent representation for faster, higher-quality and more scalable synthesis of both objects \cite{wu2024direct3d,roessle2024l3dg} and large-scale scenes \cite{zhang2023copilot4d,xiong2023learning,ran2024towards,hu2025rangeldm,ren2024xcube,meng2024lt3sd}.
However, existing works primarily focus on synthetic objects \cite{chen2023single,meng2024lt3sd} or static scenes \cite{xiong2023learning,ran2024towards,hu2025rangeldm,ren2024xcube} and cannot handle complex real-world scenarios.
In contrast, we propose a latent diffusion 3D generative model with compositional radiance fields that learns directly from in-the-wild dynamic scenes, enabling high-quality 3D asset generation at scale.

\section{Method}
\label{sec:method}

\begin{figure*}[ht]
	\centering
	\includegraphics[width=1.0\linewidth]{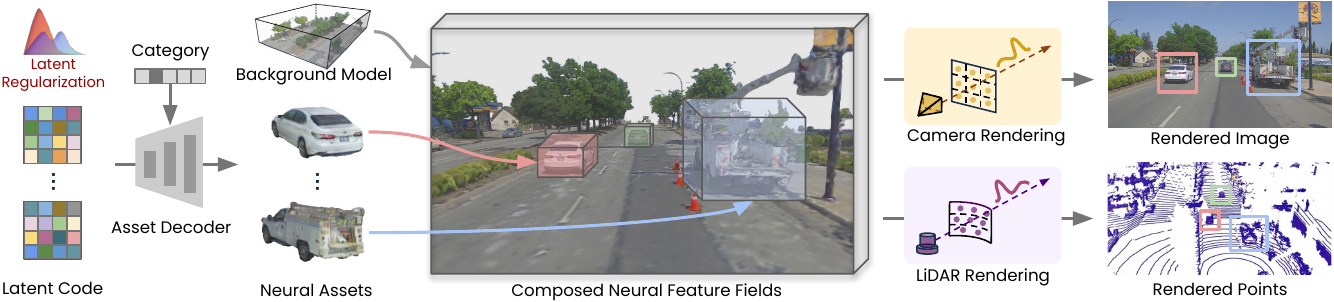}
	\vspace{-20pt}
	\caption{\textbf{Learning latent asset representation.}
		We learn a low-dimensional object latent space that generates complete assets by training across multiple scenes via occlusion-aware neural rendering.
		The asset decoder is trained to map low-dimension latent codes into neural assets which are then composed with learnable per-scene background models to match real-world sensor observations.
	}
	\vspace{-10pt}
	\label{fig:overview_recon}
\end{figure*}

\begin{figure*}[ht]
	\centering
	\includegraphics[width=1.0\linewidth]{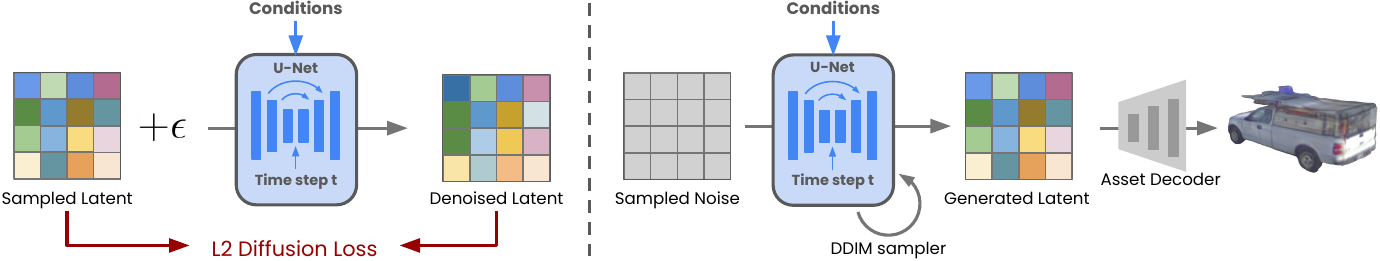}
	\vspace{-15pt}
	\caption{\textbf{Left:} Training asset diffusion model in latent space. \textbf{Right:} Sampling diffusion model for (un)conditional neural asset generation.}
	\vspace{-10pt}
	\label{fig:overview_diffusion}
\end{figure*}

Our goal is to build a generative model that can create high-fidelity assets scalably for self-driving simulation.
The model should enable unconditional generation of diverse assets for a variety of classes (e.g., cars, buses, trucks, pedestrians), and also support generating complete assets conditioned on new in-the-wild camera images and LiDAR point clouds for actors that may be occluded or far-away.
Importantly, to avoid the quality degradation \cite{chen2023single} that occurs when generative models trained on synthetic data are transferred to real data, our generative model should directly learn over a corpus of real-world driving scenes collected by a robot sensor platform.
Towards this goal, we propose a two-stage ``reconstruct-then-generate'' framework.
In the first stage, we jointly learn a set of latent codes through reconstruction-based neural rendering, where each latent code represents a foreground actor in our dataset.
We devise a compositional scene representation that enables rendering of actors of driving scenes in an occlusion-aware manner, enabling us to decouple the foreground actors from the background (\eg, road, sky, vegetation).
\cref{fig:overview_recon} provides an overview of this process.
In the second stage, we train a diffusion model to learn generative priors in this asset latent space, enabling the generation of realistic and diverse neural assets.
This enables both unconditional generation and conditional generation through score-based guidance.
\cref{fig:overview_diffusion} shows the overview of our latent asset diffusion model.
We first introduce our compositional scene representation (\cref{sec:scene_rep}).
We then describe how we compress the actor representations into a latent space to learn the set of latent codes via neural rendering (\cref{sec:latent_learning}), and detail the diffusion process in the latent space and the training process (\cref{sec:diffusion}).

\subsection{Compositional Scene Representation}
\label{sec:scene_rep}
We aim to learn a generative 3D asset model from real-world driving scenes.
Unlike existing object-centric approaches \cite{chan2022efficient, gao2022get3d, muller2023diffrf, chen2023single, gao2024cat3d, wu2024direct3d, ntavelis2023autodecoding}, our sensor data contains many actors captured under different viewpoint, distances, and occlusions.
To handle multiple instances and their occlusions, we leverage a compositional scene representation.
Each scene is decomposed into a static background $\mathcal B$ and a set of dynamic actors $\{\mathcal A_i\}_{i=1}^K$, with each actor enclosed within a 3D box and localized by a trajectory of $SE(3)$ poses.
We model the static background and dynamic actors with separate neural feature fields parameterized by triplane representations \cite{chan2022efficient}.
Specifically, for each dynamic actor's volume, explicit features are aligned along three orthogonal planes, each with resolution $N_{\mathcal A} \times N_{\mathcal A} \times C$, where $N_{\mathcal A}$ is the spatial resolution, and $C$ is the feature dimension.
Similarly, triplane features of resolution $N_{\mathcal B} \times N_{\mathcal B} \times C$ represent the background volume.
For a spatial query coordinate $\mathbf x \in \mathbb R^3$, we project it onto each of the three feature planes $\mathbf t = (\mathbf t^{xy}, \mathbf t^{xz}, \mathbf t^{yz})$ and bilinearly interpolate the corresponding feature vectors.
The interpolated features are concatenated with view direction $\mathbf d \in \mathbb S^2$ and processed by an MLP network $f_\text{feat}$ to yield geometry as a signed-distance function (SDF) $s$ and a neural feature $\mathbf f$.
This querying process is defined by:
\begin{align}
	s, \mathbf f = f_\text{feat}( \{ \texttt{interp} (\mathbf x^p, \mathbf t^p) \}_{p \in \{xy, xz, yz\} }, \mathbf d ),
	\label{eqn:nff}
\end{align}
where $\mathbf x^p$ represents the 2D projection of $\mathbf x$ onto feature plane $p$.
To render the scene representation, we first transform the object-centric actor neural feature fields to world coordinates similar to~\cite{unisim,tonderski2024neurad}.
Then we composite the background and actor feature fields.
The composited geometry and appearance features are then rendered into sensor observations through neural volume rendering.

\subsection{Learning Latent Asset Representations}
\label{sec:latent_learning}

\paragraph{Encoding to a Latent Space:}
Per-scene reconstruction methods \cite{unisim,tonderski2024neurad,yan2024street} train each scene separately, with its own set of actor and background triplanes.
This prevents effective handling of ambiguities in occluded or unseen regions, resulting in actor representations that generalize poorly to unseen viewpoints.
Naively training jointly over all the explicit triplanes for every actor in the dataset containing tens of thousands of objects requires vast amounts of GPU memory and still does not force the learned triplanes to have complete actor shape and appearance.
To address these limitations, we propose learning a latent code $\mathbf c_i \in \mathbb R^{n_{\mathcal A} \times n_{\mathcal A} \times c}$ for each actor representation and a class embedding $\mathbf e_i \in \mathbb R^{n_{\mathcal A} \times n_{\mathcal A} \times c}$, combined with a shared asset triplane decoder $f_\text{dec}$ that maps the latent code into the triplane representation:
\begin{align}
	f_\text{dec} : \mathbf e_i, \mathbf c_i \in \mathbb R^{n_{\mathcal A} \times n_{\mathcal A} \times c} \rightarrow \mathbf t_i \in \mathbb R^{N_{\mathcal A} \times N_{\mathcal A} \times 3C}.
\end{align}
The class embedding $\mathbf e_i$ has the same spatial resolution as the latent code and is shared among actors of the same class. We concatenate it with the actor latent code before passing it to the decoder.
The decoder $f_\text{dec}$ upsamples the latent code by a factor $f = N_{\mathcal A} / n_{\mathcal A}$.
The intuition is that different actors are observed from various viewpoints, so their feature planes $\mathbf t_i$ capture different informative regions.
By compressing them into a latent code bottleneck, we enable learning shape and appearance priors, allowing inference of invisible parts from sparse observations.

\paragraph{Rendering Sensor Observations:}
Now that we have decoded the asset neural fields $\{\mathbf t_i\}_{i=1}^N$ from the latent codes $\{\mathbf c_i\}_{i=1}^N$, the next step is to composite them with the background neural fields $\mathbf t_{\mathbf B}$ and render into the sensor data.
In this work, we focus on camera images and LiDAR point clouds, as they are primary sensory modalities for SDVs.

For camera rendering, we use a hybrid volume and neural rendering framework for both efficiency and realism.
Given a camera ray $\mathbf r(t) = \mathbf o + h \mathbf d$ sent from the camera center $\mathbf o$ to the pixel in direction $\mathbf d$, we sample 3D points along the ray, querying geometry $s_t$ and neural feature $\mathbf f_t$ via \cref{eqn:nff}.
We then aggregate these samples to obtain the pixel feature through volume rendering:
\begin{align}
	\mathbf f(\mathbf r) = \sum_{t=1}^{N_t} w_t \mathbf f_t, \quad w_i = \alpha_t \prod_{j=1}^{t-1} (1-\alpha_j),
	\label{eqn:camera_render}
\end{align}
where $\alpha_t$ is the opacity for the $t$-th point, derived from the SDF $s_t$ following \cite{unisim,tonderski2024neurad}.
We repeat this process to volume render all camera pixels, generating a 2D feature map $\mathbf F \in \mathbb R^{H_f \times W_f \times C_f}$.
Then we leverage 2D CNN network $f_\text{rgb}$ \cite{unisim} to convert this feature map into an image $\mathbf I_\text{rgb}$:
\begin{align}
	f_\text{rgb} : \mathbf F \in \mathbb R^{H_f \times W_f \times C_f} \rightarrow \mathbf I_\text{rgb} \in \mathbb R^{H \times W \times 3}.
\end{align}
The CNN network $f_\text{rgb}$ upsamples the rendered map from resolution $H_f \times W_f$ to $H \times W$.
This allows us to significantly reduce the number of neural feature field queries.
This approach also enhances model capacity and improves image quality by capturing spatial relationships effectively.

In addition to camera images, we use LiDAR rendering for additional geometry supervision.
LiDAR sensors emit laser pulses and measure the time of flight to determine distances to reflective surfaces.
This depth information provides valuable supervision for asset geometry.
Using similar notation, we define $\mathbf r(t) = \mathbf o + h \mathbf d$ as a ray cast from the LiDAR sensor center $\mathbf o$ in the direction $\mathbf d$.
We render LiDAR depth similarly to \cref{eqn:camera_render} by aggregating sample depths via volume rendering:
$D(\mathbf r) = \sum_{t=1}^{N_t} w_t  h_t$,
where $h_t$ is the depth of the $t$-th sampled point, and $w_t$ is the sample weight computed as in \cref{eqn:camera_render}.

\paragraph{Learning:}
We jointly optimize the asset code $\{\mathbf c_i\}_{i=1}^{N_c}$, class embedding $\{\mathbf e_i\}_{i=1}^{N_e}$, background neural fields $\mathbf t_{\mathcal B}$, asset decoder $f_\text{dec}$, neural feature fields MLP $f_\text{feat}$, and RGB CNN network $f_\text{rgb}$ by minimizing the differences between rendered and observed sensor data.
To achieve high-fidelity reconstruction and rendering, we incorporate perceptual and patch-based adversarial objectives.
We also regularize the latent space by applying a
Kullback--Leibler (KL) penalty.
Our full objective is:
\begin{align}
	\mathcal L = \mathcal L_\text{rgb}\!+\!\lambda_\text{perp} \mathcal L_\text{perp}\!+\!\lambda_\text{adv}
	\mathcal L_\text{adv}\!+\!\lambda_\text{lid} \mathcal L_\text{lid}\!+\!\lambda_\text{KL} \mathcal L_\text{KL},
\end{align}
where $L_\text{rgb}$ and $L_\text{perp}$ represent the $\ell_2$ photometric loss and the perceptual loss \cite{zhang2018unreasonable} between rendered and observed images,
$\mathcal L_\text{adv}$ is the adversarial objective using patchGAN \cite{isola2017image},
$\mathcal L_\text{lid}$ is the $\ell_1$ depth loss between rendered and observed LiDAR points,
and $\mathcal L_\text{KL}$ is the injected KL penalty guiding the latent space towards a standard normal distribution, similar to \cite{kingma2013auto,rombach2022high}:
$\mathcal L_\text{KL} = \frac{1}{2} \lVert \boldsymbol \mu_i^2 + \boldsymbol \sigma_i^2 - 1 - \log (\boldsymbol \sigma_i^2) \rVert_1$,
where $\boldsymbol \mu_i$ and $\boldsymbol \sigma_i$ represent the mean and standard deviation components of latent code $\mathbf c_i$, \ie, $\mathbf c_i = \boldsymbol \mu_i^2 + \boldsymbol \sigma_i \odot \boldsymbol \epsilon$, with $\boldsymbol \epsilon \sim \mathcal N(\mathbf 0, \mathbf I)$.
This regularization prevents overfitting and encourages the latent space to be compact, continuous, smooth, and low-variance, enabling easier learning of generative models.
Similar to \cite{xu2023discoscene,gadde2021detail}, we introduce additional perceptual losses on object patches for enhanced object-level supervision, where the patches are obtained by projecting the 3D instance boxes onto the image plane.
Please refer to supp. for more details.

\subsection{Latent Asset Diffusion Model}
\label{sec:diffusion}
Given our learned asset code library $\{\mathbf c_i\}_{i=1}^N$ from the dataset, we aim to learn generative priors using a diffusion model.
Diffusion models \cite{sohl2015deep,ho2020denoising} are probabilistic frameworks that model the data distribution by progressively denoising a Gaussian noise variable.
In our approach, the diffusion model operates directly in the latent space and begins with Gaussian noise, progressively denoising it to recover the underlying latent distribution.
The \textit{forward} or \textit{diffusion} process is a discrete-time Markov chain that iteratively adds Gaussian noise to the latent code $\mathbf c^{(0)}$, producing progressively noisier codes following the transition probability $q(\mathbf c^{(t)} | \mathbf c^{(t-1)}) = \mathcal N( \sqrt{1-\beta^{(t)}} \mathbf c^{(t-1)}, \beta^{(t)} \mathbf I )$, according to a noise schedule $\beta^{(t)}$.
After $T$ steps, the code approximates pure Gaussian noise.
This \textit{forward} process can be directly sampled at timestep $t$ using closed-form expression:
\begin{align}
	\mathbf c^{(t)} = \sqrt{\bar\alpha^{(t)}} \mathbf c^{(0)} + \sqrt{1 - \bar\alpha^{(t)}} \boldsymbol \epsilon,
	\label{eqn:forward_process}
\end{align}
where $\bar\alpha^{(t)} = \prod_{s=1}^t \alpha^{(s)}$ with $\alpha^{(t)} = 1-\beta^{(t)}$, and $\boldsymbol \epsilon \sim \mathcal N (\mathbf 0, \mathbf I)$ is drawn from a Gaussian distribution.
The goal of the diffusion model is to learn the \textit{reverse} process, progressively denoising the noisy code to recover a clean version.
We parameterize the denoising network $f_\text{diff}$ as a U-Net \cite{ronneberger2015u}, which takes the noisy code $\mathbf c^{(t)}$ and timestep $t$ as inputs, predicting the noise estimate $f_\text{diff} (\mathbf c^{(t)}, t)$ and removing it to yield a denoised code.
The denoising network is trained with a weighted $\ell_2$ objective:
\begin{align}
	\mathcal L_\text{diff} = \mathbb E_{\mathbf c, \boldsymbol \epsilon, t} \left[ \frac{1}{2} w^{(t)} \lVert f_\text{diff} (\mathbf c^{(t)}, t) - \boldsymbol \epsilon \rVert_2^2 \right],
\end{align}
where $\boldsymbol \epsilon \sim \mathcal N (\mathbf 0, \mathbf I)$ is standard Gaussian noise,
$t \sim \mathcal U(1, T)$ is uniformly sampled from the interval $[1, T]$,
and $\mathbf c^{(t)}$ is derived from $\mathbf c^{(0)}$ (\cref{eqn:forward_process}), where $\mathbf c^{(0)} \sim \{\mathbf c_i\}_{i=1}^N$ is sampled from the asset code library.
The term $w^{(t)}$ provides time-dependent weighting.
Learning the diffusion process in the latent space, rather than in high-dimensional triplane space like prior work \cite{muller2023diffrf,chen2023single,shue20233d,bautista2022gaudi,wu2024direct3d}, offers key advantages for likelihood-based generative modeling by:
(i) focusing on essential contents of the data and
(ii) operating in a computationally efficient, compact space.

\paragraph{Unconditional Asset Generation:}
Unconditional asset generation involves generating a latent code using the learned diffusion model $f_\text{diff}$ and then decoding it into neural assets using the learned asset decoder $f_\text{dec}$.
Sampling from the diffusion prior can be performed with various solvers (\eg, DDPM \cite{ho2020denoising}, DDIM \cite{song2020denoising}), similar to methods used in image generation.
Taking DDIM \cite{song2020denoising} for example, we begin with a random Gaussian code $\mathbf c^{(T)} \sim \mathcal N (\mathbf 0, \mathbf I)$ and iteratively denoise it over $T$ steps until reaching $t = 0$:
\begin{align}
	\label{eqn:uncond_gen}
	\mathbf c^{(t-1)}                           & = \sqrt{\alpha^{(t-1)}}
	\left(
	\frac{\mathbf c^{(t)} - \sqrt{\beta^{(t)}} \tilde{\boldsymbol \epsilon}}{\sqrt{\alpha^{(t)}}}
	\right)
	+
	\sqrt{\beta^{(t-1)}} \tilde{\boldsymbol \epsilon},
	\\
	\text{s.t. }~~ \tilde{\boldsymbol \epsilon} & = f_\text{diff}(\mathbf c^{(t)}, t),
\end{align}
where $\tilde{\boldsymbol \epsilon}$ is the noise estimate at timestep $t$ from the denoising network $f_\text{diff}$.
The final denoised code $\mathbf c^{(0)}$ is a sample from the learned latent distribution.
We then pass this code to the asset decoder to generate the asset neural fields, $\mathbf t = f_\text{dec}(\mathbf c^{(0)})$, for scene composition and rendering.

\paragraph{Conditional / Guided Assets Generation:}
For conditional generation, we aim to model the conditional distribution of the latent space $p(\mathbf c | y)$.
This can be achieved with a conditional denoising network $f_\text{diff} (\mathbf c^{(t)}, t, y)$, with condition signals such as actor class and time-of-day.
As shown in Fig.~\ref{fig:teaser}, classifier-free diffusion guidance enables effective control over asset generation.
Alternatively, for some inverse problems such as image-to-3D, conditional generation can be approximated from an unconditional model using classifier guidance \cite{dhariwal2021diffusion,song2020score}, leveraging gradients of the rendering loss \wrt known observations.
This flexibility removes the need for separate models tailored to different inverse problems.
Given camera image and/or LiDAR point cloud of a new actor to reconstruct, during the sampling process (\cref{eqn:uncond_gen}) we compute the gradient of the rendering loss with respect to the current code:
$\mathbf g = \nabla_{\mathbf c^{(t)}} (\mathcal L_\text{rgb} + \lambda_\text{lid} \mathcal L_\text{lid})$,
and then steer the sampling process by updating the noisy code similar to~\cite{muller2023diffrf,chen2023single}:
\begin{align}
	\label{eqn:cond_gen}
	\hat{\mathbf c}^{(t-1)} \leftarrow \mathbf c^{(t-1)} -\lambda \cdot \mathbf g,
\end{align}
where $\mathbf c^{(t-1)}$ is computed following \cref{eqn:uncond_gen} and $\hat{\mathbf c}^{(t-1)}$ is the updated code; $\lambda$ is a small guidance weight.

\section{Experiments}
\label{sec:experiments}

\begin{figure*}[ht]
	\centering
	\includegraphics[width=1.0\linewidth]{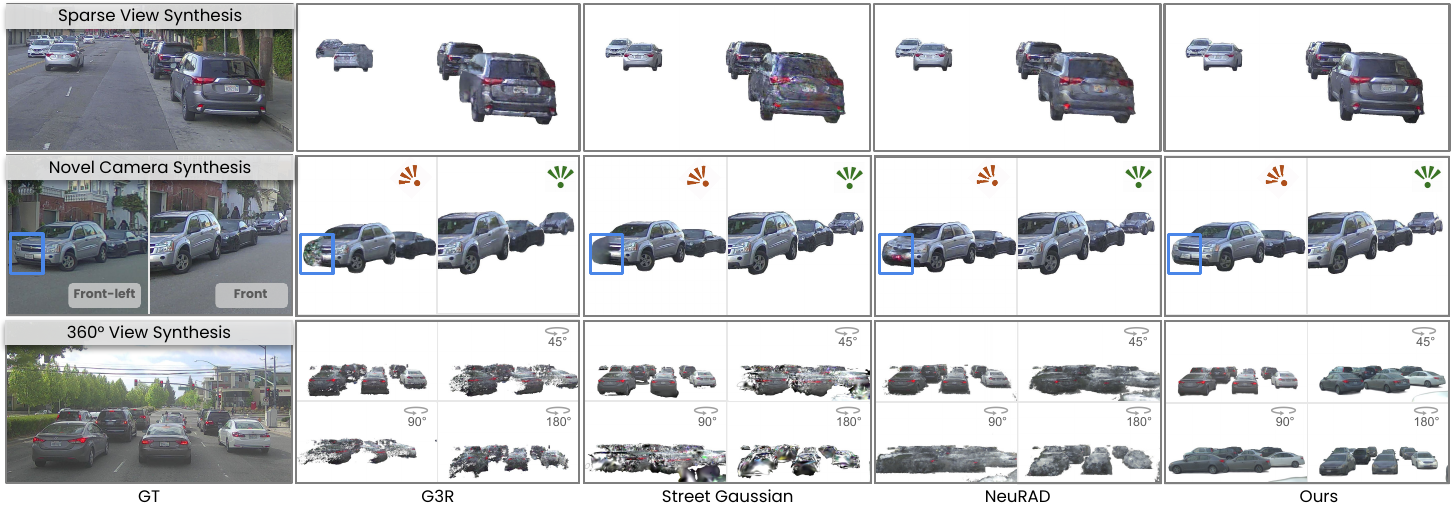}
	\vspace{-15pt}
	\caption{
		\textbf{Top: Sparse view synthesis.}
		\methodname{} generalizes well on this extreme setting thanks to low-dimensional latent space learned across many scenes, while the SoTA reconstruction methods are less robust and produce noticeable visual artifacts (\eg, missing, blurry or distorted appearance).
		\textbf{Middle: Novel camera synthesis.}
		We train on frames from the front camera and evaluate on frames from the front-left camera.
		Our method generates assets that have better extrapolation results, while retaining high quality in the original view.
		\textbf{Bottom: $360^\circ$ view synthesis.}
		Through our multi-scene training and latent space, our method enables higher-fidelity asset completion for different object orientations compared to SoTA reconstruction-based methods.
	}
	\label{fig:qual_recon}
\end{figure*}

\begin{table*}[ht]
	\centering
	\begin{tabular}{lccccccc}
		\toprule
		\multirow{2}{*}{Methods}                              & \multicolumn{3}{c}{Sparse View Synthesis} &
		\multicolumn{3}{c}{Novel Camera Synthesis}            & \multicolumn{1}{c}{360° View Synthesis}
		\\
		\cmidrule(l){2-4} \cmidrule(l){5-7} \cmidrule(l){8-8} &
		{PSNR$\uparrow$ }                                     & {SSIM$\uparrow$ }                         & {LPIPS$\downarrow$ } &
		{PSNR$\uparrow$ }                                     & {SSIM$\uparrow$ }                         & {LPIPS$\downarrow$ } &
		FID$\downarrow$
		\\
		\midrule
		PixelSplat*~\cite{charatan2024pixelsplat}             & 17.67                                     & 0.704                & 0.336          & -              & -              & -              & -
		\\
		G3R~\cite{chen2025g3r}                                & 18.37                                     & 0.711                & 0.255          & 17.40          & 0.723          & 0.221          & 191.92
		\\
		\midrule
		Street Gaussian~\cite{yan2024street}                  & 20.01                                     & 0.763                & 0.156          & 17.81          & 0.724          & 0.226          & 162.37
		\\
		NeuRAD~\cite{tonderski2024neurad}                     & 21.07                                     & \textbf{0.825}       & 0.129          & 17.49          & 0.723          & 0.214          & 159.38
		\\
		\midrule
		Ours                                                  & \textbf{21.34}                            & \textbf{0.825}       & \textbf{0.113} & \textbf{18.36} & \textbf{0.805} & \textbf{0.147} & \textbf{100.28}
		\\
		\bottomrule
	\end{tabular}
	\caption{
		\textbf{Quantitative  comparison with SoTA reconstruction approaches on PandaSet.}
		We evaluate on sparse view synthesis (10\% frames for training and remaining frames for testing), novel camera synthesis (training on front camera and testing on front-left camera), and $360^\circ$ view synthesis (rotating actors from $0^\circ$ to $360^\circ$).
		\methodname{} outperforms existing SoTA per-scene or generalizable reconstruction approaches across all settings.
		PixelSplat* \cite{charatan2024pixelsplat} (2 input views) only for sparse view synthesis evaluation.
	}
	\label{tab:recon_comparison}
\end{table*}

In this section, we first describe our experimental setup.
We then compare our model to state-of-the-art reconstruction methods across multiple novel view synthesis settings.
Next, we compare to SoTA generative models for 3D generation.
Finally, we demonstrate conditional generation from different classes, single-image, and appearance (day/night), and show that the generated assets enhance downstream detection performance.
Please refer to the supp. for implementation details and ablation of our model components.

\subsection{Experimental Setup}
\label{sec:setup}

\paragraph{Dataset:}
We conduct experiments on the PandaSet \cite{xiao2021pandaset} dataset, which includes $103$ driving scenes captured in San Francisco, with each scene spanning $8$ seconds ($80$ frames at $10$hz).
The data collection platform features a $360^\circ$ mechanical spinning LiDAR, a forward-facing LiDAR, along with six $1920 \times 1080$ cameras surrounding the vehicles.
Object bounding boxes are provided with the dataset.
As our focus is object reconstruction and generation, we extract object segmentation masks for metric evaluation.
Specifically, we leverage 3D object boxes and LiDAR points to identify actor patches in the camera images, and then apply a visual foundation model \cite{kirillov2023segment} to generate instance masks for the actors.
Our method does not require instance masks for learning.
We employ them strictly to evaluate foreground-only metrics.
Please see supp. for details.

\paragraph{Baselines:}
We compare our method with several state-of-the-art reconstruction and generation approaches.
For reconstruction, we benchmark against per-scene NeRF-based \textit{NeuRAD} \cite{tonderski2024neurad} and 3D Gaussian Splatting-based \textit{Street Gaussians} \cite{yan2024street}, as well as generalizable reconstruction methods \textit{G3R} \cite{chen2025g3r} and \textit{PixelSplat} \cite{charatan2024pixelsplat}.
During evaluation, these methods take as input all source views for the full scene, except for PixelSplat, which can only handle two views.
For generation, we evaluate against the GAN-based \textit{EG3D} \cite{chan2022efficient} and \textit{DiscoScene} \cite{xu2023discoscene}, as well as diffusion-based \textit{SSDNeRF} \cite{chen2023single}.
As \textit{EG3D} and \textit{SSDNeRF} are object-centric models, we train them on our object-centric benchmark with instance masks.

\subsection{Reconstruction Evaluation}
\label{sec:recon_eval}
\paragraph{Evaluation Settings and Metrics:}
We evaluate our method across three challenging settings:
(1) \textit{Sparse view synthesis:} Using every $10$th frame for training and the remaining frames for testing, with both training and testing frames captured from the front camera.
(2) \textit{Novel camera synthesis:} Training on frames from the front camera and evaluating on frames from the front-left camera.
(3) \textit{$360^\circ$ View synthesis:} Rotating actors ($0^\circ$–$360^\circ$) to simulate various behaviors, evaluated on front camera views.
We select $7$ diverse scenes to assess reconstruction performance, with the remaining $96$ scenes for learning generalizable priors.
For \textit{sparse view synthesis} and \textit{novel camera synthesis}, we report PSNR, SSIM \cite{wang2004image}, and LPIPS \cite{zhang2018unreasonable}.
For \textit{$360^\circ$ view synthesis}, since no ground truth images are available, we report FID metrics \cite{heusel2017gans}.
As PixelSplat \cite{charatan2024pixelsplat} is designed to reconstruct with two input views, we only evaluate it on the \textit{sparse view synthesis} setting.

\paragraph{Sparse View Synthesis:}
We first compare our method against state-of-the-art approaches for sparse view synthesis in \cref{tab:recon_comparison}.
Our method outperforms the baselines across all metrics.
Unlike prior methods that uses $90\%$ frames \cite{chen2024omnire,yang2023emernerf}, $75\%$ frames \cite{yan2024street} or $50\%$ frames \cite{unisim,tonderski2024neurad,chen2025g3r} for training, our setup is more challenging, relying on only $10\%$ of frames for training.
We found that baseline methods struggle to learn robust geometry, leading to severe artifacts in novel viewpoints.
Qualitative results in \cref{fig:qual_recon} (top) show that our method renders camera images more accurately.

\paragraph{Novel Camera Synthesis:}
We compare our method against SoTA for novel camera synthesis in \cref{tab:recon_comparison}.
A visual comparison is provided in \cref{fig:qual_recon} (middle).
All baseline methods effectively reconstruct observed camera data (front camera).
However, for new sensor poses (front-left camera), previously unobserved regions become visible, requiring the model to ``hallucinate" these areas.
For instance, the front-right part of the SUV in \cref{fig:qual_recon} (middle) is invisible in the front camera but becomes visible in front-left camera views.
Reconstruction methods fail on these regions.
Our method, by learning generative priors, is capable of rendering complete and high-quality simulations in this setting.

\paragraph{$\mathbf{360}^\circ$ View Synthesis:}
Sensor simulation requires not only simulating novel viewpoints but also synthesizing entirely new scenarios.
Consider a four-way intersection with a car crossing in front of you from a perpendicular lane.
To explore a situation where that car suddenly turns sharply into your lane, we must render actors in significantly different poses than those seen in the original scene.
To that end, we investigate a $360^\circ$ view synthesis setting by rotating actors in the scene from $0^\circ$ to $360^\circ$, simulating various orientations.
Since ground-truth images are unavailable, we report FID \cite{heusel2017gans} in \cref{tab:recon_comparison}.
Qualitative results in \cref{fig:qual_recon} (bottom) show that our method effectively hallucinates unseen parts of the actors, whereas all baselines struggle to do so.

\begin{figure*}[ht]
	\centering
	\includegraphics[width=1.0\linewidth]{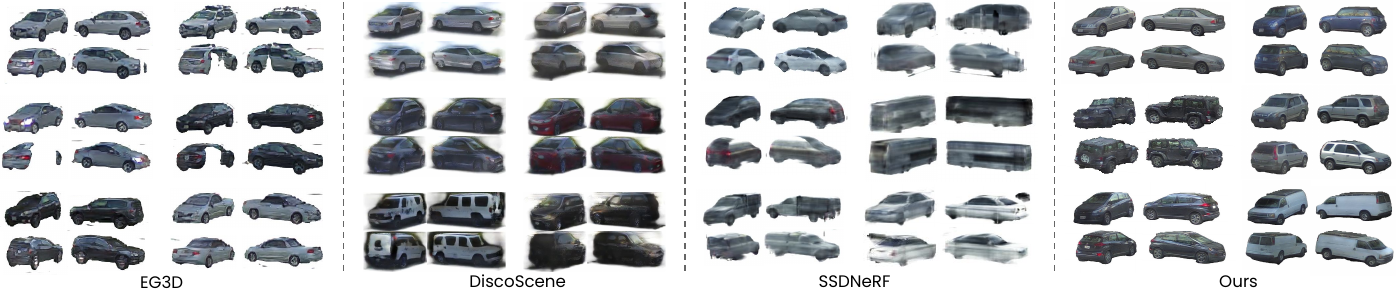}
	\vspace{-20pt}
	\caption{
		\textbf{Qualitative comparison on unconditional generation.}
		Our methods generates more diverse, complete and higher-quality 3D assets compared to SoTA 3D generative models.
	}
	\vspace{-15pt}
	\label{fig:qual_uncond_gen}
\end{figure*}

\begin{figure}[ht]
	\centering
	\includegraphics[width=0.9\linewidth]{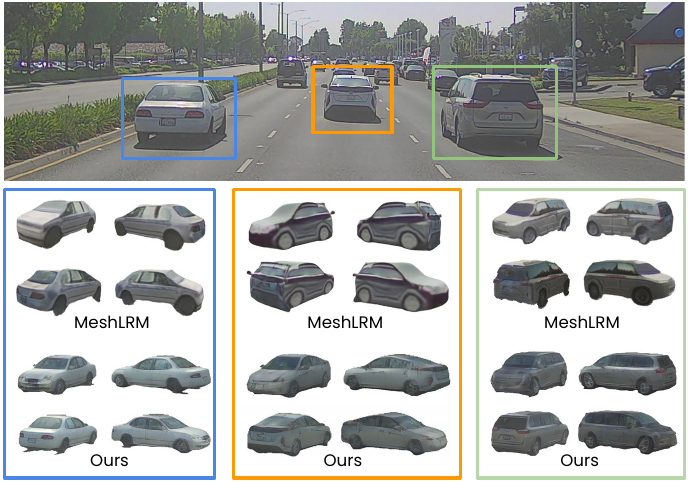}
	\vspace{-5pt}
	\caption{\textbf{Qualitative results on single-image to 3D.}}
	\vspace{-12pt}
	\label{fig:qual_cond_gen}
\end{figure}

\subsection{Generation Evaluation}
\label{sec:gen_eval}

\paragraph{Evaluation Metrics:}
To evaluate the generation quality, we report both Fréchet Inception Distance (FID) \cite{heusel2017gans} and Kernel Inception Distance (KID) \cite{binkowski2018demystifying} scores.
The metrics are calculated by comparing distributions of $10$K generated images with all available images in the validation set, providing a quantitative assessment of similarity in terms of perceptual quality and realism.
FID measures the distance between the mean and covariance of feature representations from real and generated image distributions.
KID complements FID by measuring differences in feature embeddings through a kernel-based approach.
All generated objects are rendered at a resolution of $256 \times 256$.

\vspace{-5pt}
\paragraph{Unconditional Generation:}
We compare our method against SoTA GAN-based method EG3D \cite{chan2022efficient} and DiscoScene \cite{xu2023discoscene}, as well as diffusion-based method SSDNeRF \cite{chen2023single} in \cref{tab:generation_comparison}.
All baselines struggle with occlusion, leading to suboptimal assets.
In addition, EG3D and DiscoScene suffers from mode collapse, producing limited asset variety (mainly sedans).
DiscoScene struggles to differentiate foreground and background and exhibits boundary artifacts, likely due to the absence of LiDAR or mask supervision.
SSDNeRF yields blurry results, potentially due to challenges posed by our real-world sparse setting.
Our model improves over these baselines across all metrics.
\cref{fig:qual_uncond_gen} shows a visual comparison.

\subsection{Applications}
\label{sec:application}

\paragraph{Conditional Generation:}
The flexibility of our framework enables various conditional generation tasks.
Specifically, we freeze the learned latent codes and train a conditional diffusion model $f_\text{diff} (\mathbf c^{(t)}, t, y)$ using classifier-free guidance.
We explore conditioning on fine-grained actor classes and time-of-day (day/night), with results presented in \cref{fig:teaser}.
We can generate complete assets for a variety classes, with intra-class shape and appearance variation.

\vspace{-10pt}
\paragraph{Single Image to 3D:}
Our approach also enables generating 3D assets from single-view images using a rendering-guided denoising process (\cref{eqn:cond_gen}), where the denoising gradient is directed to minimize rendering loss against the observed image.
\cref{fig:qual_cond_gen} shows qualitative results of single-image reconstruction for three nearby vehicles, alongside comparisons to SoTA large reconstruction model MeshLRM~\cite{wei2024meshlrm}.
MeshLRM produces distorted shapes and synthetic cartoon-ish appearance for unobserved views.
Our approach generates higher quality completion and is consistent with the input.
Please refer to supp. for more comparisons with other SoTA large models.

\vspace{-10pt}
\paragraph{Data Augmentation with \methodname{}:}
We now showcase that our generated assets boost downstream performance when training a BEVFormer \cite{li2022bevformer} 3D object detector (\cref{tab:data_aug}).
Specifically, we augment the training dataset by swapping out existing actors with generated ones for the same scene layouts.
Please see supp. for more details.

\begin{table}[t]
	\centering
	\begin{tabular}{lrr}
		\toprule
		Methods                            & FID$\downarrow$ & KID$\downarrow$
		\\
		\midrule
		SSDNeRF \cite{chen2023single}      & 191.30          & 156.13
		\\
		DiscoScene \cite{xu2023discoscene} & 138.48          & 116.27
		\\
		EG3D \cite{chan2022efficient}      & 80.56           & 38.97
		\\
		\midrule
		Ours                               & \textbf{59.50}  & \textbf{28.32}
		\\
		\bottomrule
	\end{tabular}
	\vspace{-5pt}
	\caption{
		\textbf{Unconditional generation evaluation on PandaSet.}
		KID scores are multiplied by $10^3$.
		\methodname{} outperforms SoTA baselines on both metrics and generates higher-quality 3D assets.
	}
	\label{tab:generation_comparison}
\end{table}

\begin{table}[t]
	\centering
	\resizebox{0.48\textwidth}{!}{
		\begin{tabular}{lcccc}
			\toprule
			           & mAP$\uparrow$  & AP@1m$\uparrow$ & AP@2m$\uparrow$ & AP@4m$\uparrow$ \\
			\midrule
			Real       & 27.08          & 8.58            & 26.99           & 45.67           \\
			Real + Sim & \textbf{29.32} & \textbf{9.78}   & \textbf{29.18}  & \textbf{49.00}  \\
			\bottomrule
		\end{tabular}
	}
	\vspace{-5pt}
	\caption{\textbf{Data augmentation with \methodname{} helps 3D detection.} We report distance-based APs~\cite{caesar2020nuscenes} at $1m$, $2m$ and $4m$.}
	\label{tab:data_aug}
	\vspace{-5pt}
\end{table}
\section{Conclusion}
\label{sec:conclusion}

In this work, we tackled the challenge of generating high-quality and complete assets from in-the-wild LiDAR and camera data captured by a moving sensor platform.
Towards this goal, we developed a ``reconstruct-then-generate'' approach where we first learn to reconstruct foreground actors over multiple scenes with compositional scene neural rendering and encode them to a latent space.
We then train a diffusion model to operate within this latent space to enable generation.
We show our method generates high-quality, complete assets for actors such as vehicles and motorcycles, outperforming both per-scene reconstruction methods and generative models.
We also show that our approach can be conditioned for controllable asset generation such as on sparse sensor data, actor class, and time of day, enabling diverse and scalable content creation for simulation.
Future work involves generating dynamic lighting and animation, intrinsic decomposition, and adopting more efficient scene representation and rendering techniques.

\vspace{-10pt}

\paragraph{Acknowledgement:}
We sincerely thank the Waabi team for their invaluable assistance and support.
We also thank the anonymous reviewers for their insightful suggestions.

{
    \small
    \bibliographystyle{ieeenat_fullname}
    \bibliography{main}
}

\clearpage

\onecolumn
{\hspace{-5mm} \LARGE{\textbf{Supplementary Material}}\\}
\maketitle
\appendix
\appendix
\renewcommand{\thetable}{A\arabic{table}}
\setcounter{table}{0}
\renewcommand{\thefigure}{A\arabic{figure}}
\setcounter{figure}{0}

In the supplementary material, we provide implementation details of our method (\cref{sec:additional_details}) and baselines (\cref{sec:baseline_details}), experimental settings (\cref{sec:exp_details}), additional results and analysis (\cref{sec:additional_results}), and then discuss limitations and future directions of our method (\cref{sec:limitation}).
Please refer to our project page \url{https://waabi.ai/genassets} for an overview of our methodology and video results of \methodname{} for scalable content creation and sensor simulation.

\section{\methodname{} Details}
\label{sec:additional_details}

\paragraph{Scene Representation Model:}
Our scene representation model is based on tri-plane feature maps and neural feature fields MLP network $f_\text{feat}$.
For each actor, we use a tri-plane with spatial resolution of $128 \times 128$ and feature dimension of $12$.
For the background, we use a triplane with spatial resolution $256 \times 256$ and feature dimension of $12$.
The neural feature fields MLP network is composed of two sub-networks.
The first one takes the interpolated feature $\{\texttt{interp} (\mathbf x^p, \mathbf t^p)\}_{p \in \{xy,xz,yz\}}$ as input and predicts the signed distance value $s$ and an intermediate feature.
The second network takes the concatenation of the intermediate feature and viewpoint encoding as input to predict the neural feature vector $\mathbf f$ with $32$ channels.
To model the unbounded scene, we adopted an inverted sphere parameterization similar to \cite{unisim, barron2022mipnerf360}.
To decouple shadows from actors, we identify the camera ray's intersection with the actor's bottom plane (derived from the actor's pose and dimension) and use a shadow head MLP to predict the actor's shadow RGB.

\paragraph{Rendering Details:}
To render the neural feature fields, we sample query points with a step size of $5$ cm for regions inside the foreground actor bounding boxes, and $30$ cm otherwise.
To enable efficient volume rendering for unbounded background regions, we leverage geometry priors from LiDAR observations to localize near-surface regions, restricting radiance field evaluations to these areas.
This approach significantly reduces the number of samples and radiance queries needed.
Specifically, we construct an occupancy grid for the scene volume based on aggregated LiDAR point clouds similar to \cite{unisim}, with a voxel size of $0.5$ m.
Additionally, $8$ extra points are sampled for the distant sky region during volume rendering.

\paragraph{Asset Decoder:}
Our asset decoder $f_\text{dec}$ is designed to transform the asset latent representation into the tri-plane representation.
The latent code has a spatial resolution of $8 \times 8$ and a feature dimension of $32$.
We regularize the latent space with a small Kullback-Leibler (KL) penalty using the reparameterization trick \cite{kingma2013auto}.
The asset decoder adopts a similar architecture to \cite{esser2021taming}, where the latent code is concatenated with the class embedding and processed through a sequence of blocks: a residual block with $256$ channels, a non-local block with $256$ channels, and another residual block with $256$ channels.
This is followed by five residual blocks with channel channels $256, 128, 128, 64, 64$, with four $2\times$upsampling layers interleaved from the second to the last residual block, progressively increasing the spatial resolution to $128 \times 128$.
Finally, a convolutional layer predicts the tri-plane representation.
Group normalization with $32$ groups is applied to normalize intermediate features.

\paragraph{Camera RGB CNN Network:}
The camera RGB network $f_\text{rgb}$ consists of $6$ residual blocks with $32$ channels, and upsample the rendered image feature map from $480 \times 270$ to $1920 \times 1080$ resolution.
A convolution layer is applied at the beginning to convert input feature to $32$ channels, and another convolution layer is applied to predict the final output image.
To get a larger receptive field, we set kernel size to 5 for all residual blocks.
Two $2\times$ upsampling layers are inserted after the second residual block and the fourth residual block.

\paragraph{Denoising Network:}
The denoising network $f_\text{diff}$ is implemented as a U-Net \cite{ronneberger2015u}, following the design in DDPM \cite{ho2020denoising}.
The U-Net uses a base channel of $128$ and includes two $2\times$ scaling module, each consisting of two residual blocks.
The feature dimension increases progressively from the base channel size of $128$ to $512$.
Timestep embeddings $t$ and conditioning signals (\eg, class labels, time-of-day) are incorporated into the intermediate layers.
For the prediction format, we use the $v$-parameterization, as proposed in \cite{salimans2022progressive}.

\paragraph{Training Details:}
In the first stage, we jointly train the asset code $\{\mathbf c_i\}_{i=1}^{N_c}$, class embedding $\{\mathbf e_i\}_{i=1}^{N_e}$, background neural fields $\mathbf t_{\mathcal B}$, asset decoder $f_\text{dec}$, neural feature fields MLP $f_\text{feat}$, and RGB CNN network $f_\text{rgb}$ to minimize the reconstruction objective (Eqn. 5 in the main paper).
The training set consists of $6$ cameras from PandaSet across $96$ training logs, each containing $80$ images at $1920 \times 1080$ resolution, totaling approximately $46$k images.
Training is performed on $16$ NVIDIA A10G GPUs, for $100$ epochs, taking approximately $1.5$ days to complete.
The loss weights in the learning objective are set as follows:
$\lambda_\text{rgb}=1$,
$\lambda_\text{perp}=0.1$,
$\lambda_\text{adv}=0.001$,
$\lambda_\text{lid}=0.01$,
and $\lambda_\text{KL}=1e-5$.
For the additional perceptual loss on object patches, we project the 3D instance boxes onto images to extract patches, which are resized to $256 \times 256$ resolution for perceptual loss computation.
We adopt Adam optimizer \cite{kingma2014adam} for training.
Learning rate is $0.05$ for asset latent code, $0.0001$ for asset decoder and class embedding, $0.001$ for neural feature fields network, and $0.001$ for RGB CNN network.
In the second stage, the denoising network is trained exclusively on $8$ NVIDIA A10G GPUs for $100$k iterations, requiring approximately $12$ hours.
The Adam optimizer is used with a learning rate of $0.0001$ for the denoising network.

\section{Baseline Implementation Details}
\label{sec:baseline_details}

\subsection{Reconstruction Baselines}

\paragraph{NeuRAD \cite{tonderski2024neurad}:} Following \cite{unisim}, NeuRAD leverages compositional neural radiance fields to handle dynamic scenes. It further proposes several techniques to handle more complex sensor phenonmena (\eg, rolling-shutter, ray-dropping and beam divergence) and achieves superior performance in camera simulation. We adopt the public implementation\footnote{\url{https://github.com/georghess/neurad-studio}} and train the models for 20,000 iterations. The models are trained on all input views (9 images for sparse view synthesis, and 80 images for novel camera synthesis and 360$^\circ$ view synthesis) for each validation snippet.

\paragraph{Street Gaussian \cite{yan2024street}:} Street Gaussian replaces NeRFs in \cite{unisim,tonderski2024neurad} with compositional 3DGS and achieves real-time camera simulation, but does not support LiDAR. We adopt the public implementation\footnote{\url{https://github.com/ziyc/drivestudio}} and train the models for 30,000 iterations with the default hyperparameters (\eg, density control, learning rate schedule). We use 800,000 downsampled aggregated LiDAR points and random 200,000 points for the initialization of 3D Gaussians. We train the models on all input views (9 images for sparse view synthesis, and 80 images for novel camera synthesis and 360$^\circ$ view synthesis) for each validation snippet.

\paragraph{PixelSplat \cite{charatan2024pixelsplat}:} PixelSplat is a generalizable scene reconstruction approach based on 3D Gaussian Splatting. It predicts 3D Gaussians with a 2-view epipolar transformer to extract features and then predicts the depth distribution and pixel-aligned Gaussians. The model is trained with 96 training PandaSet snippets with 9 source views and 71 target views. During sparse view synthesis evaluation, we select the two nearest source views to predict and render the representation.
We adopt the public implementation\footnote{\url{https://github.com/dcharatan/pixelsplat}} and use $2\times$ A6000 (48GB) to train the models.
Due to GPU memory constraints, we downscale the image
resolution to 360 $\times$ 640 for PandaSet (and rescale images to 1920 $\times$ 1080 for evaluation).
We note that the original work uses an 80GB A100 for training and handles $256 \times 256$ resolution. We use \texttt{re10k} config and train each model for 100k iterations with a batch size of 1.

\paragraph{G3R \cite{chen2025g3r}:} G3R is a generalizable reconstruction approach that is designed for real-world large scenes and can efficiently predict high-quality 3D Gaussians taking many training images. It learns a reconstruction network that takes the gradient feedback signals from differentiable rendering to iteratively update a 3D scene representation. We adapt G3R to our evaluation setting by training models on sparse training views (sparse view synthesis) or all front-camera images (novel camera synthesis and 360$^\circ$ view synthesis).
For all experiments, during training we follow \cite{chen2025g3r} to randomly select 20 consecutive frames (10 source views, 10 target views) using the front camera.
We train the models on 2$\times$ A6000 (48GB) for 300 epochs.
During the evaluation, we take all source images to reconstruct the 3D representations (9 images for sparse view synthesis, 80 images for novel camera synthesis and 360$^\circ$ view synthesis) for each validation snippet, and then render at the target views.

\subsection{Generation Baselines}

\paragraph{EG3D \cite{chan2022efficient}:}
EG3D is a GAN-based model for high-quality 3D object synthesis with implicit representations.
It combines StyleGAN-like latent space manipulation with a geometry-aware approach, leveraging a tri-plane representation to balance rendering quality and computational efficiency.
The architecture integrates volumetric rendering and an efficient super resolution module, enabling fast and scalable training while maintaining photorealistic results and precise 3D structures.
We adopt the official implementation\footnote{\url{https://github.com/NVlabs/eg3d}} and train the models on our object-centric benchmark ($\approx 55$k object images) using $2\times$ A5000 (24GB) GPUs.
We render the feature map at a resolution of $64 \times 64$ and use $4\times$ super resolution module to produce final images at $256 \times 256$ resolution.
Each ray samples $64$ coarse samples and $64$ importance samples.
The model is trained with a batch size of $8$ images over a total of $1000$k images.
We apply $R_1$ regularization with $\gamma=1$ to ensure stable training.

\paragraph{DiscoScene \cite{xu2023discoscene}:}
DiscoScene is a 3D-aware GAN-based generative model for controllable scene synthesis.
It utilizes an abstract object-level representation as the scene layout prior and spatially disentangles the scene into object-centric generative radiance fields by learning solely from 2D images with the global-local discriminators.
We use the official implementation\footnote{\url{https://github.com/NVlabs/eg3d}} and follow the provided configuration\footnote{\url{https://github.com/snap-research/discoscene/blob/master/scripts/discoscene/training/train_waymo256.sh}} for Waymo.
We train the model on all cameras from PandaSet, with a heuristic-based filtering process to remove noisy object samples.
Specifically, we exclude objects smaller than $200$ pixels in the original $1920 \times 1080$ resolution or those occluded by closer objects (\ie, having $>50\%$ IoU with a nearer object when projecting the 3D instance box onto the camera image).
We train the model using $2\times$ A5000 (24GB) GPUs for $300$k iterations.

\paragraph{SSDNeRF \cite{chen2023single}:}
SSDNeRF (Single-Stage Diffusion Neural Radiance Field) is a SoTA diffusion-based model for high-quality 3D object generation.
It proposes a single-stage training paradigm with an end-to-end objective that jointly optimizes the tri-plane NeRF reconstruction and a diffusion model in the triplane space.
This design enables simultaneous 3D reconstruction and generative prior learning.
SSDNeRF achieves competitive results in both unconditional generation and sparse-view 3D reconstruction tasks.
We adopt the official implementation\footnote{\url{https://github.com/Lakonik/SSDNeRF}} and use $2\times$ A5000 (24GB) GPUs to train the models on our object-centric benchmark.
We follow the config for unconditional generation detailed in the official repo\footnote{\url{https://github.com/Lakonik/SSDNeRF/blob/main/configs/paper_cfgs/ssdnerf_cars_uncond.py}}.
We train the model with rendered images at a resolution of $256 \times 256$ and tri-plane resolution of $128 \times 128$.
Since instances in our PandaSet object-centric benchmark contain varying numbers of images, we randomly sample three views per instance in each iteration.
The model is trained with a batch size of $16$ instances over $400$k iterations.

\section{Experiment Details}
\label{sec:exp_details}

\subsection{Pandaset Dataset}
\begin{figure*}[t]
	\centering
	\includegraphics[width=1.0\linewidth]{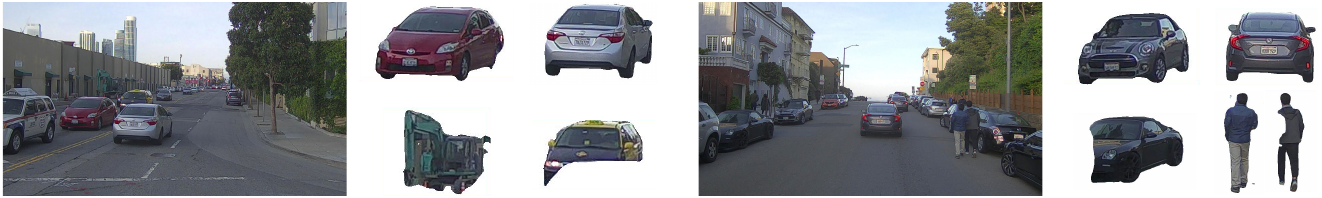}
	\caption{
		\textbf{Examples of object-centric benchmark.}
		We leverage Segment Anything \cite{kirillov2023segment} model and object 3D bounding boxes to create an object-centric benchmark for evaluating object reconstruction and generation, and training object-centric baselines \cite{chan2022efficient,chen2023single}.
	}
	\label{fig:supp_dataset}
\end{figure*}

We evaluate on public real-world dataset PandaSet \cite{xiao2021pandaset} which contains $103$ urban driving scenes captured in San Francisco.
Each scene spans $8$ seconds ($80$ frames sampled at $10$Hz).
The data collection platform consists of a $360^\circ$ mechanical spinning LiDAR as well as a forward-facing LiDAR, along with $6$ HD ($1920 \times 1080$) cameras.
These cameras are facing front, front-left, left, back, front-right, and right.
PandaSet also provides human annotated 3D bounding boxes in each frame.
Following \cite{chen2025g3r}, we select $7$ diverse scenes (\texttt{001, 030, 040, 080, 090, 110, 120}) for evaluation, and the remaining $96$ scenes for training.
As our focus is object reconstruction and generation, we leverage Segment Anything model (SAM) \cite{kirillov2023segment} to extract object segmentation masks for metric evaluation.
We also use them to train object-centric baselines EG3D \cite{chan2022efficient} and SSDNeRF \cite{chen2023single}.
Specifically, we project 3D bounding boxes onto the camera image to obtain 2D boxes.
The 2D bounding boxes are used as prompts for the SAM, which generates per-pixel object masks.
Next, we project the 3D LiDAR points within the 3D bounding box onto the camera image and calculate the ratio of points falling outside the object mask.
To handle failures in SAM that are inconsistent with the 3D annotation, instances with more than $30\%$ of LiDAR points falling outside the masks are removed.
We also filter out instances that are too small ($<100$ pixels) in the camera images.
To further improve the data quality for training object-centric baselines, we exclude instances that are heavily occluded by other objects.
Specifically, we project actor bounding boxes onto the camera image and determine the occluded areas caused by closer actor bounding boxes.
Instances with more than $50\%$ of their bounding box area occluded are removed.
After processing, we have around 55k object images (with corresponding instance masks) to train the object-centric baselines and around 5k images for evaluation.
\cref{fig:supp_dataset} shows examples of object-centric images created from PandaSet.

\begin{table*}[t]
	\centering
	\begin{tabular}{llll}
		\toprule
		\methodname{} Actor Class   &  &  & PandaSet Actor Class                 \\
		\midrule
		Car                         &  &  & Car
		\\
		\midrule
		\multirow{3}{*}{Truck}      &  &  & Pickup Truck
		\\
		                            &  &  & Medium-sized Truck
		\\
		                            &  &  & Semi-truck
		\\
		\midrule
		Bus                         &  &  & Bus
		\\
		\midrule
		Construction Vehicle        &  &  & Other Vehicle - Construction Vehicle
		\\
		\midrule
		Emergency Vehicle           &  &  & Emergency Vehicle
		\\
		\midrule
		Bicycle                     &  &  & Bicycle
		\\
		\midrule
		\multirow{2}{*}{Motorcycle} &  &  & Motorcycle
		\\
		                            &  &  & Motorized Scooter
		\\
		\midrule
		\multirow{2}{*}{Pedestrian} &  &  & Pedestrian
		\\
		                            &  &  & Pedestrian with Object
		\\
		\bottomrule
	\end{tabular}
	\caption{
		Mappings between PandaSet actor class and our defined actor class for class conditional generation.
	}
	\label{tab:supp_class_mapping}
\end{table*}

\subsection{Evaluation Metric Details}

\paragraph{Asset Reconstruction Metrics:}
For \textit{Sparse view synthesis} and \textit{Novel camera synthesis} settings, we report peak signal-to-noise ratio (PSNR), structural similarity (SSIM) \cite{wang2004image}, and perceptual similarity (LPIPS) \cite{zhang2018unreasonable} metrics to evaluate the photorealism of novel views.
We mask out the background using our inferred object segmentation masks.
Our approach, along with generalizable reconstruction baselines PixelSplat \cite{charatan2024pixelsplat} and G3R \cite{chen2023single}, are trained on the $96$ training logs to learn generalizable priors.
For \textit{Sparse view synthesis}, we evaluate using the front camera from $7$ evaluation logs.
All methods take every $10$th frame as source frames and evaluate on the remaining frames.
For \textit{Novel camera synthesis}, we use all frames from the front camera as source frames and evaluate on all frames from the front-left camera across the $7$ evaluation logs.
For \textit{$360^\circ$ View synthesis}, we rotate actors by $0^\circ, 45^\circ, 90^\circ, 135^\circ, 180^\circ, 225^\circ, 270^\circ, 315^\circ$ to simulate various behaviors to evaluate the asset completeness.
All frames from the front camera are used as source frames, and evaluations are conducted on the $7$ evaluation logs.
Since ground-truth images for rotated actors are unavailable, we measure Fréchet Inception Distance (FID) \cite{heusel2017gans} between the rotated images and source images.

\paragraph{Asset Generation Metrics:}
To evaluate the asset generation quality, we report both the Fréchet Inception Distance (FID) \cite{heusel2017gans} and Kernel Inception Distance (KID) \cite{binkowski2018demystifying} metrics.
The FID metric is defined as:
\begin{align}
	\mathrm{FID} = \lVert \mu_r - \mu_g \rVert^2 + \mathrm{Tr} \left( \Sigma_r + \Sigma_g - 2 (\Sigma_r  \Sigma_g)^\frac{1}{2} \right)
\end{align}
where $\mu_r$ and $\mu_g$ are the means of the feature representations for the real and generated images, respectively.
And $\Sigma_r$ and $\Sigma_g$ are the corresponding covariance matrices.
$\mathrm{Tr}$ denotes the trace of a matrix.
The KID metric measures the squared Maximum Mean Discrepancy (MMD) between two distributions using a polynomial kernel.
For both FID and KID, we generate $10$k object images and compute the metrics \wrt all object images in the evaluation set.
We render the images at a resolution of $256 \times 256$, and resize the real images to the same resolution.

\begin{figure*}[t]
	\centering
	\includegraphics[width=1.0\linewidth]{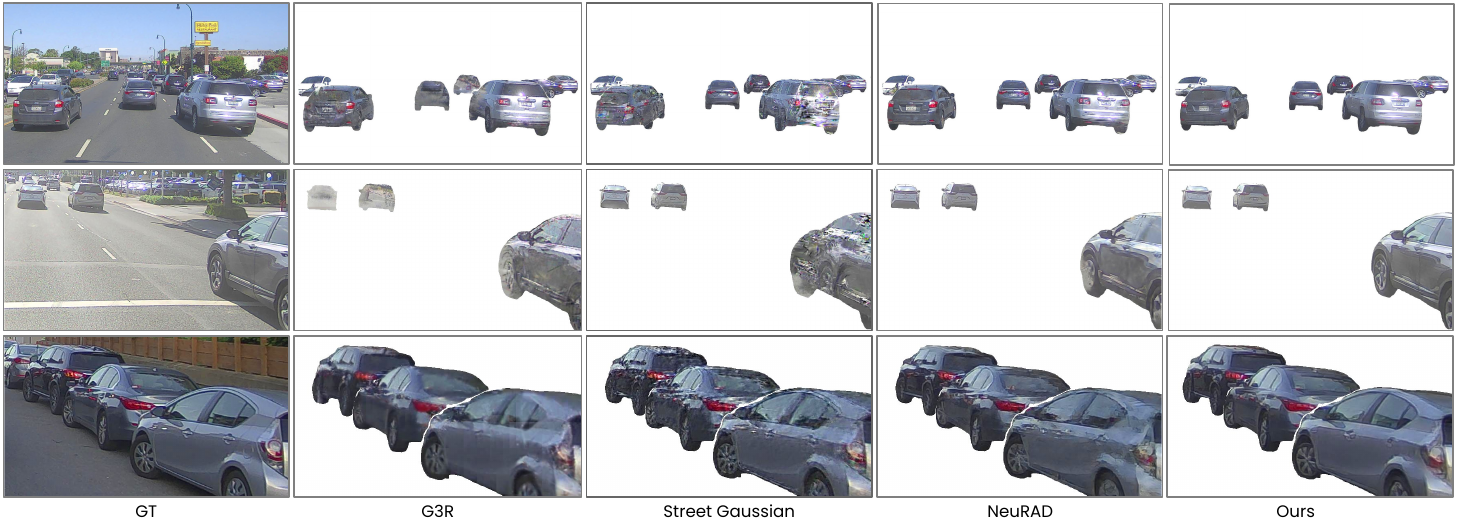}
	\caption{
		\textbf{Qualitative results on sparse view synthesis.}
		All methods use every $10$th frame as source frames and evaluate on the remaining frames.
		\methodname{} generalizes well under this challenging setting, whereas SoTA reconstruction methods show reduced robustness and introduce noticeable visual artifacts.
	}
	\label{fig:supp_recon_sparse}
\end{figure*}

\begin{figure*}[t]
	\centering
	\includegraphics[width=1.0\linewidth]{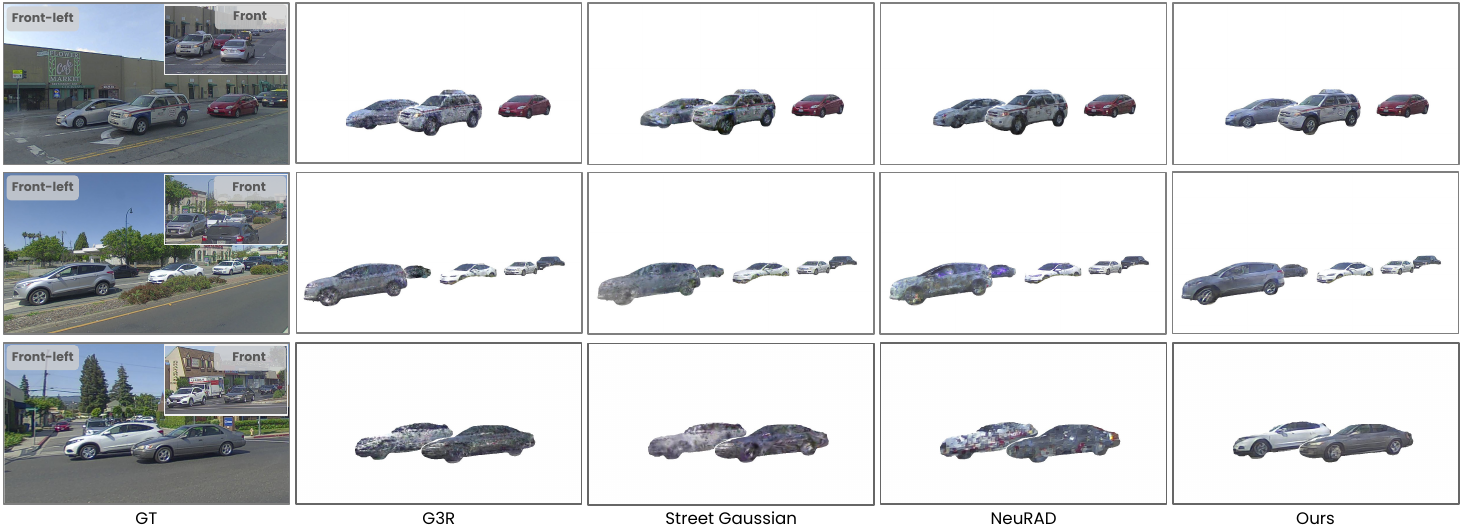}
	\caption{
		\textbf{Qualitative results on novel camera synthesis.}
		We train on front-camera frames and evaluate on front-left-camera frames.
		\methodname{} achieves better extrapolation, while baseline methods exhibit significant visual artifacts.
	}
	\label{fig:supp_recon_novel_sensor}
\end{figure*}

\begin{figure*}[t]
	\centering
	\includegraphics[width=1.0\linewidth]{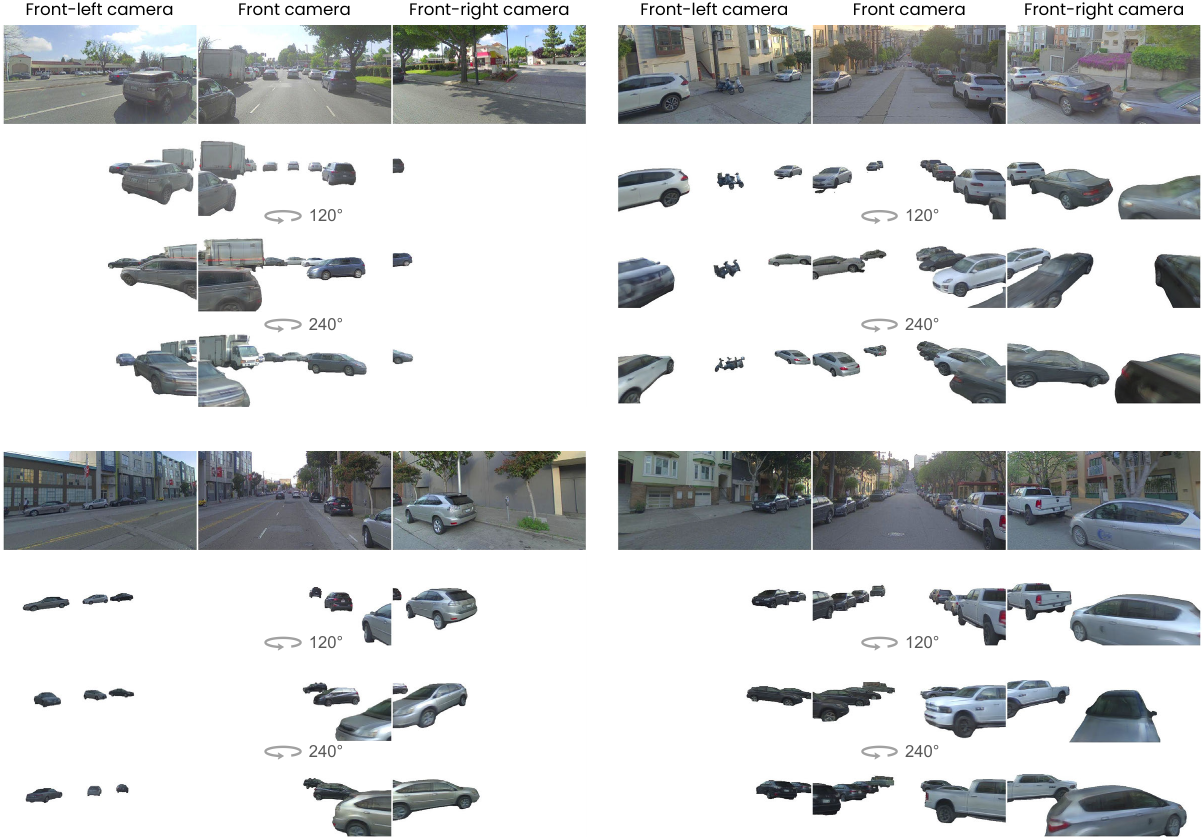}
	\caption{
		\textbf{Qualitative results on $360^\circ$ view synthesis.}
		Leveraging multi-scene training and a structured latent space, \methodname{} enables higher-fidelity asset completion across $360^\circ$ orientations.
	}
	\label{fig:supp_recon_360}
\end{figure*}

\begin{figure*}[t]
	\centering
	\includegraphics[width=\linewidth]{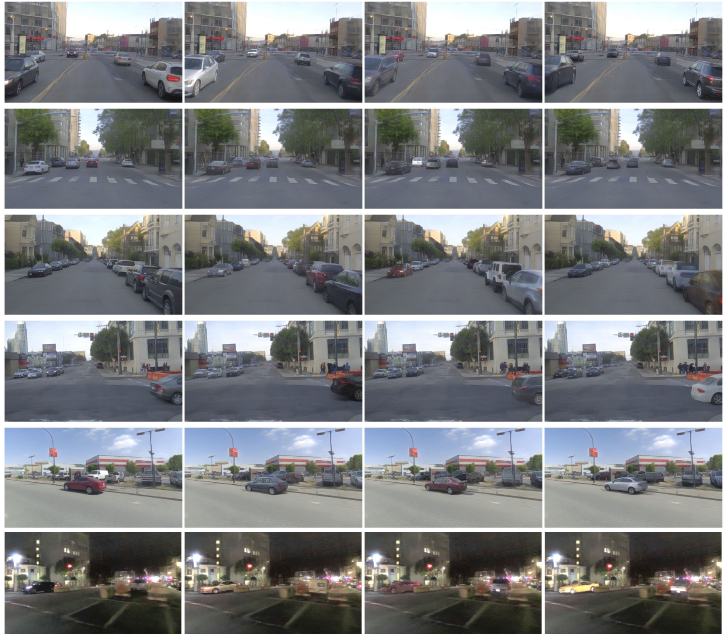}
	\caption{
		\textbf{Examples of sensor simulation generated by \methodname{}.}
		For each scene layout (row), existing actors are replaced with generated ones, showcasing different variations per layout.
	}
	\label{fig:supp_augmentation}
\end{figure*}

\begin{figure*}[t]
	\centering
	\includegraphics[width=1.0\linewidth]{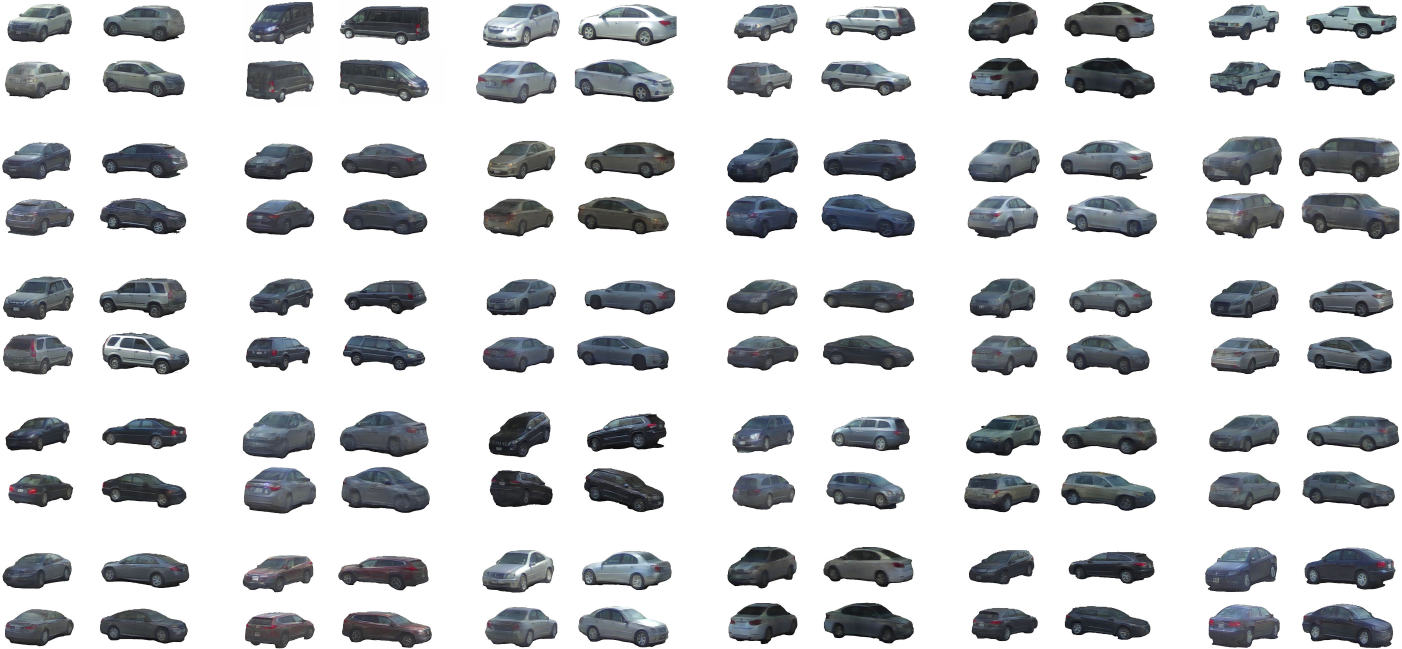}
	\caption{
		\textbf{Additional qualitative results on unconditional generation.}
	}
	\label{fig:supp_uncond_gen}
\end{figure*}

\begin{figure*}[t]
	\centering
	\includegraphics[width=1.0\linewidth]{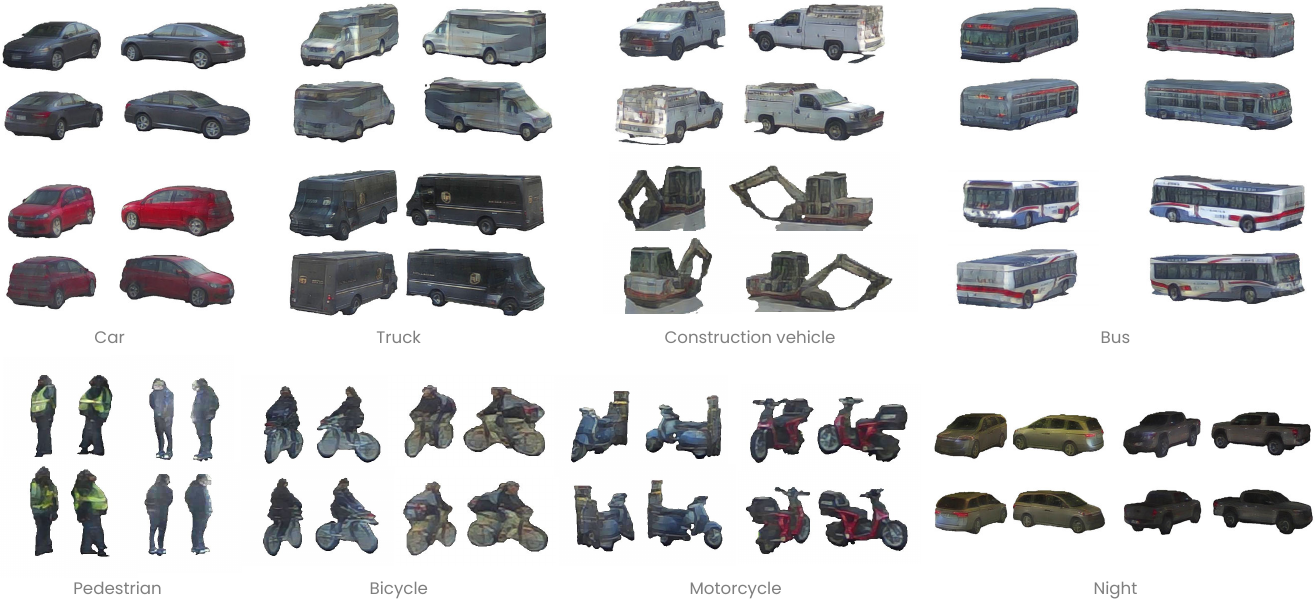}
	\caption{
		\textbf{Additional qualitative results on conditional generation.}
	}
	\label{fig:supp_cond_gen}
\end{figure*}

\begin{figure*}[t]
	\centering
	\includegraphics[width=\linewidth]{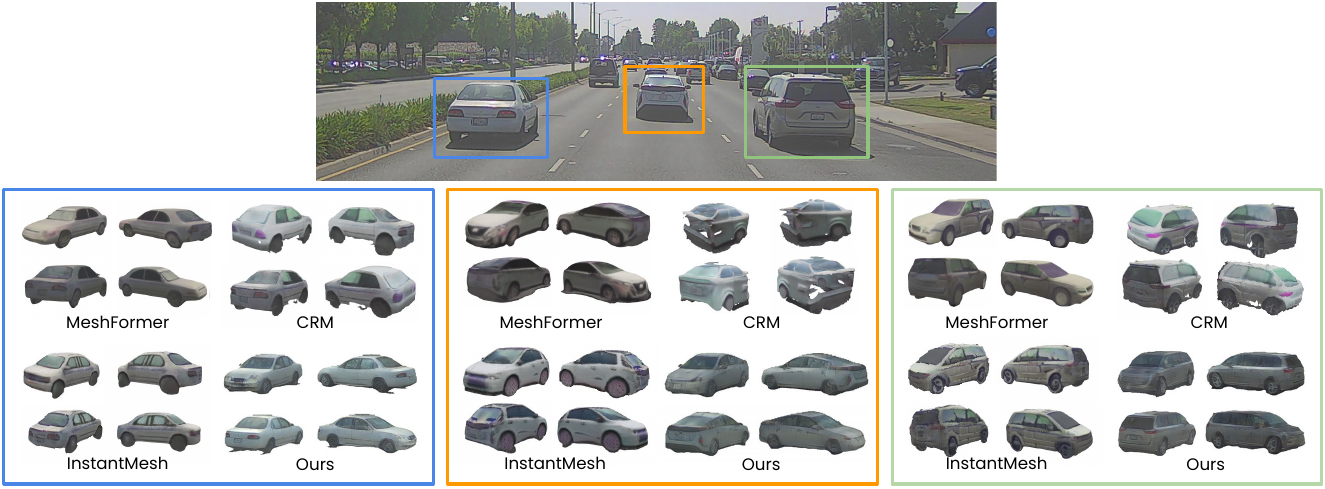}
	\caption{
		\textbf{Additional qualitative results on single image to 3D.}
	}
	\label{fig:supp_cond_recon}
\end{figure*}

\subsection{\methodname{} Experiments Details}

\paragraph{Asset Reconstruction Details:}
We train \methodname{} on $96$ training logs and fine-tune on the $7$ evaluation logs for evaluating asset reconstruction metrics.
During fine-tuning, we optimize the asset latent code and background tri-planes to minimize the reconstruction objective (Eqn. 5 in the main paper).
In addition, we incorporate a diffusion prior term (Eqn. 7 in the main paper), resulting in the final training objective: $\mathcal L = \mathcal L_\text{rend} + \lambda_\text{diff} \mathcal L_\text{diff}$.
By guiding the gradient to minimize the diffusion loss, we achieve a more complete reconstruction of the actor latent code.
This approach, known as score distillation sampling (SDS) \cite{poole2022dreamfusion}, is widely used in radiance fields generation.

\paragraph{Asset Generation Details:}
We train \methodname{} on $96$ training logs, which include diverse assets classes and different times of day.
For class-conditional generation, we map the PandaSet classes into $8$ fine-grained \methodname{} classes according to ~\cref{tab:supp_class_mapping}.
We use an \texttt{Embedding} layer to encode the one-hot representation of class information, which is then combined with the timestep embedding before being fed into the U-Net denoising network.
For time-of-day conditional generation, actors from night logs (\texttt{057, 058, 059, 062, 063, 064, 065, 066, 067, 068, 069, 070, 071, 072, 073, 074, 077, 078, 079, 149}) are classified as night actors, while the others are categorized as day actors.
The time-of-day information is then used to condition the diffusion model in the same way as the class-conditional model.
For single-image-to-3D generation, we employ a rendering-guided denoising process (Eqn. 10 in main paper).
We jointly generate multiple actors within the same image and applying the rendering loss to the composited actor representations, which helps to mitigate the shape ambiguity caused by occlusions.

\paragraph{Data Augmentation Details:}
We show that our generated assets boost downstream performance when training the BEVFormer detector.
Specifically, we augment the training dataset by swapping out existing actors with generated ones for the same scene layouts.
\cref{fig:supp_augmentation} shows examples of the augmented dataset.
We adapt the official BEVFormer repository\footnote{\url{https://github.com/fundamentalvision/BEVFormer}} to support PandaSet, focusing on single-frame vehicle detection using only the front camera. Any actor outside the camera’s field of view is ignored. Our models are trained in vehicle coordinates following the FLU convention (x: forward, y: left, z: up), with the region of interest defined as .
$x\in[0,80m]$,$y\in[-40m,40m]$, and $z\in[-2m,6m]$.
We use the BEVFormer-tiny architecture with a batch size of 4 per GPU. We use the AdamW optimizer and employ the default cosine learning rate schedule for 25 epochs.
For augmented training, we generate a simulated (sim) dataset by replacing existing actors with generated ones while keeping the same scene layouts, and then mix this with the original data (real).
We only consider the vehicle class and report distance-based AP \cite{caesar2020nuscenes} at $1m, 2m$, and $4m$, and average the three metrics to obtain the final mAP.

\section{Additional Experiments and Analysis}
\label{sec:additional_results}

\subsection{Additional Qualitative Results on Reconstruction}

\paragraph{Sparse View Synthesis:}
Additional qualitative comparisons of sparse view synthesis are presented in \cref{fig:supp_recon_sparse}.
\methodname{} demonstrates strong generalization in this challenging setting, while SoTA reconstruction methods exhibit reduced robustness and introduce noticeable visual artifacts.

\paragraph{Novel Camera Synthesis:}
We compare \methodname{} against SoTA methods for novel camera synthesis in \cref{fig:supp_recon_novel_sensor}.
The inset shows source images from the front camera.
Rendering from a new camera (front-left) involves significant viewpoint changes, with previously unobserved regions becoming visible.
\methodname{} handles these challenges effectively, whereas baselines struggle with robustness and produce visible artifacts.

\paragraph{$360^\circ$ View synthesis:}
Additional $360^\circ$ view synthesis results are shown in \cref{fig:supp_recon_360}.
\methodname{} captures complete asset shapes and appearances, enabling high-fidelity rendering across $360^\circ$ viewpoints, allowing for scalable sensor simulation.

\subsection{Additional Qualitative Results on Generation}

\paragraph{Unconditional Generation:}
We provides additional unconditional generation results in \cref{fig:supp_uncond_gen}.
\methodname{} generates diverse, complete and higher-quality 3D assets and enables scalable content creation for sensor simulation.

\paragraph{Conditional Generation:}
\cref{fig:supp_uncond_gen} provides additional asset generation results conditioned on class and time-of-day.
\methodname{} allows us to control the generation process for diverse asset creation.

\paragraph{Single Image to 3D:}
We provides additional single image to 3D comparisons in \cref{fig:supp_cond_recon}.
Compared to SoTA 3D large  models MeshFormer \cite{liu2024meshformer}, CRM \cite{wang2025crm} and InstantMesh \cite{xu2024instantmesh},  our approach generates higher quality 360$^\circ$ completion and is more multi-view consistent. Due to the reliance on object-centric synthetic dataset training, existing 3D large models usually produce cartoonish generation results especially on unobserved views.

\begin{table*}[t]
	\centering
	\begin{tabular}{lccccccc}
		\toprule
		\multirow{2}{*}{Methods}                              & \multicolumn{3}{c}{Sparse View Synthesis}      &
		\multicolumn{3}{c}{Novel Camera Synthesis}            & \multicolumn{1}{c}{$360^\circ$ View Synthesis}
		\\
		\cmidrule(l){2-4} \cmidrule(l){5-7} \cmidrule(l){8-8} &
		{PSNR$\uparrow$ }                                     & {SSIM$\uparrow$ }                              & {LPIPS$\downarrow$ } &
		{PSNR$\uparrow$ }                                     & {SSIM$\uparrow$ }                              & {LPIPS$\downarrow$ } &
		FID$\downarrow$
		\\
		\midrule
		Full model                                            & 21.34                                          & 0.825                & 0.113          & \textbf{18.36} & \textbf{0.805} & \textbf{0.147} & \textbf{100.28}
		\\
		w/o KL regularizer                                    & 21.48                                          & 0.825                & \textbf{0.112} & 18.06          & 0.800          & 0.150          & 110.87
		\\
		w/o perceptual supervision                            & \textbf{21.52}                                 & \textbf{0.829}       & 0.155          & 18.15          & 0.803          & 0.177          & 110.19
		\\
		Tri-plane opt. w/o asset decoder                      & 19.31                                          & 0.784                & 0.186          & 17.58          & 0.778          & 0.195          & 169.44
		\\
		\bottomrule
	\end{tabular}
	\caption{
		\textbf{Ablation study on asset reconstruction.}
	}
	\label{tab:ablation_recon}
\end{table*}

\begin{table}[t]
	\centering
	\begin{tabular}{lrr}
		\toprule
		Methods            & FID$\downarrow$ & KID$\downarrow$
		\\
		\midrule
		Full model         & \textbf{59.50}  & \textbf{28.32}
		\\
		w/o KL regularizer & 82.67           & 39.06
		\\
		\bottomrule
	\end{tabular}
	\caption{
		\textbf{Ablation study on unconditional generation.}
	}
	\label{tab:ablation_gen}
\end{table}

\subsection{Ablation Study}
In this section, we study the effectiveness of several key components of \methodname{} on PandaSet.
\cref{tab:ablation_recon} reports the reconstruction metrics.
The KL term plays a important role in regularizing the latent space and learning complete asset representation, which can impacts the novel camera synthesis and $360^\circ$ view synthesis results where the viewpoint changes are significant.
The perceptual supervision enhances the overall image quality (LPIPS and FID) by preserving more details.
We also investigate a setting where per-actor triplane representation are learned directly instead of using latent codes with a shared asset decoder in latent space.
We use the same resolution of $128 \times 128$ for the per-actor triplane representation and learn them on the $7$ evaluation logs in our experiment.
This approach struggles to capture complete assets, resulting in notably worse performance particularly for $360^\circ$ view synthesis.
Finally, we study the impact of KL regularizer in diffusion learning in \cref{tab:ablation_gen}.
The KL regularizer is essential for learning complete asset representations for diffusion learning in our in-the-wild setting.

\section{Discussions}
\label{sec:limitation}

\subsection{Artifacts and potential enhancements}
Our reconstructed or generated assets are not perfect.
Our method can still lack sharp details for unobserved views and have inconsistent color tone.
Increasing model capacity and dataset size may ameliorate these artifacts.
The triplane representation can have boundary-like smearing artifacts.
Our work may benefit from enhanced 3D representations and rendering techniques \cite{barron2023zip,kerbl20233d}.
We also do not model various sensor effects (\eg, sensor calibration \cite{yang2025unical}) or label noise \cite{unisim} in the real-world data, which might result in degraded performance.

\subsection{Limitations and Future Work}
\methodname{} assumes all objects are rigid.
Despite this, we still find that the method reconstructs reasonable shape and appearance for pedestrians, as depicted in \cref{fig:supp_cond_gen} and Fig. 1 of the main paper.
Additionally, \methodname{} are not fully lighting-aware.
Future works include 4D modelling and animation \cite{chen2024omnire}, and intrinsic decomposition \cite{pun2023neural}, as well as more powerful conditioning techniques to further control content creation.

\end{document}